\documentclass[twoside]{article}
\usepackage[a4paper]{geometry}
\usepackage[latin1]{inputenc} 
\usepackage[T1]{fontenc} 
\usepackage{RR}
\usepackage{natbib}
\usepackage{hyperref}
\usepackage{xspace}
\usepackage{booktabs}
\usepackage{graphicx}

\newcommand*{\vv}[1]{\vec{\mkern0mu#1}}
\def\melt{MElt\xspace}
\def\marmot{MarMoT\xspace}
\def\lefff{\mbox{Le\textit{fff}}\xspace}
\def\leffe{\mbox{Le\textit{ff}e}\xspace}

\RRNo{8924}
\RRdate{June 2016}
\RRauthor{
Beno\^it Sagot
}
\authorhead{Sagot}
\RRtitle{Utilisation d'informations lexicales externes pour l'annotation multilingue en parties du discours}
\RRetitle{External Lexical Information for Multilingual Part-of-Speech Tagging}
\titlehead{External Lexical Information for Multilingual Part-of-Speech Tagging}
\RRresume{Les lexiques morphosyntaxiques et les représentations vectorielles des mots ont chacun montré leur utilité
  pour améliorer la précision d'étiqueteurs morphosyntaxiques statistiques. Nous comparons ici les performances de
  quatre systèmes sur des jeux de données couvrant 16 langues, deux de ces systèmes reposant sur des traits (MEMM et
  CRF) et deux autres sur des approches neuronales (bi-LSTM). Nous montrons qu'en moyenne les quatre approches
  obtiennent des performances similaires de niveau état-de-l'art. Néanmoins, nos modèles reposant sur des traits ont de
  meilleures performances sur les jeux de données lexicalement plus riches (par exemple sur des langues à morphologie
  riche), alors que les résultats obtenus par les approches neuronales sont meilleurs sur les jeux de données dont la
  variabilité lexicale est moindre (par exemple pour l'anglais). Ces conclusions sont vraies en particulier pour nos
  modèles de type MEMM faisant usage de notre système \melt, qui s'appuie sur un jeu de traits renouvelé. Ceci montre
  que, sous certaines conditions, les approches par traits enrichies par des lexiques morphosyntaxiques sont compétitifs
  par rapport aux approches neuronales.
}
\RRabstract{ Morphosyntactic lexicons and word vector representations have both proven useful for improving the accuracy
  of statistical part-of-speech taggers. Here we compare the performances of four systems on datasets covering 16
  languages, two of these systems being feature-based (MEMMs and CRFs) and two of them being neural-based (bi-LSTMs). We
  show that, on average, all four approaches perform similarly and reach state-of-the-art results.  Yet better
  performances are obtained with our feature-based models on lexically richer datasets (e.g.~for morphologically rich
  languages), whereas neural-based results are higher on datasets with less lexical variability (e.g.~for
  English). These conclusions hold in particular for the MEMM models relying on our system \melt, which benefited from
  newly designed features. This shows that, under certain conditions, feature-based approaches enriched with
  morphosyntactic lexicons are competitive with respect to neural methods.  }
\RRmotcle{\'Etiquetage en partie du discours, Modèles reposant sur des traits, Modèles neuronaux, MEMM, CRF, bi-LSTM,
  Analyse multilingue}
\RRkeyword{Part-of-Speech Tagging, Feature-based models, Neural models, MEMM, CRF, bi-LSTM, Multilingual Analysis}
\RRprojets{ALPAGE}
\RCParis 

\begin{document}
\makeRR   
\section{Introduction}

Part-of-speech tagging is now a classic task in natural language processing, for which many systems have been developed
or adapted for a large variety of languages. Its aim is to associate each ``word'' with a morphosyntactic tag, whose
granularity can range from a simple morphosyntactic category, or part-of-speech (hereafter PoS), to finer categories
enriched with morphological features (gender, number, case, tense, mood, etc.).

The use of machine learning algorithms trained on manually annotated corpora has long become the standard way to develop
PoS taggers. A large variety of algorithms have been used, such as (in approximative chronological order) bigram and
trigram hidden Markov models \citep{merialdo94,brants96,brants00}, decision trees \citep{schmid94,magerman95}, maximum
entropy Markov models (MEMMs) \citep{ratnaparkhi96} and Conditional Random Fields (CRFs)
\citep{lafferty01,constant12}. With such machine learning algorithms, it is possible to build PoS taggers for any
language, provided adequate training data is available. {}

As a complement to annotated corpora, it has previously been shown that external lexicons are valuable
sources of information, in particular morphosyntactic lexicons, which provide a large inventory of (word,~PoS)
pairs. Such lexical information can be used in the form of constraints at tagging time \citep{kim99,hajic00tagging} or
during the training process as additional features combined with standard features extracted from the training corpus
\citep{chrupala08,goldberg09,denis12}.

In recent years, a different approach to modelling lexical information and integrating it into natural language
processing systems has emerged, namely the use of vector representations for words or word sequences
\citep{bengio03,collobert08,chrupala13,ling15,ballesteros15,muller15}. Such representations, which are generally
extracted from large amounts of raw text, have proved very useful for numerous tasks including PoS tagging, in
particular when used in recurrent neural networks (RNNs) and more specifically in mono- or bi-directional, word-level
and/or character-level long short-term memory networks (LSTMs) \citep{hochreiter97,ling15,ballesteros15,plank16}.

Both approaches to representing lexical properties and to integrating them into a PoS tagger improve tagging
results. Yet they rely on resources of different natures.  The main advantage of word vectors is that they are built in
an unsupervised way, only requiring large amounts of raw textual data. They also encode finer-grained information than
usual morphosyntactic lexicons, most of which do not include any quantitative data, not even simple frequency
information. Conversely, lexical resources often provide information about scarcely attested words, for which
corpus-based approaches such as word vector representations are of limited relevance. Moreover, morphological or
morphosyntactic lexicons already exist for a number of languages, including less-resourced langauges for which it might
be difficult to obtain the large amounts of raw data necessary to extract word vector representations.

Our main goal is therefore to compare the respective impact of external lexicons and word vector representations on the
accuracy of PoS models. This question has already been investigated for 6 languages by \citet{muller15} using the
state-of-the-art CRF-based tagging system \marmot. The authors found that their best-performing word-vector-based PoS
tagging models outperform their models that rely on morphosyntactic resources (lexicons or morphological analysers). In
this paper, we report on larger comparison, carried out in a larger multilingual setting and comparing different tagging
models. Using different 16 datasets, we compare the performances of two feature-based models enriched with external
lexicons and of two LSTM-based models enriched with word vector representations. A secondary goal of our work is to
compare the relative improvements linked to the use of external lexical information in the two feature-based models,
which use different models (MEMM vs.~CRF) and feature sets.

More specifically, our starting point is the \melt system \citep{denis12}, an MEMM tagging system. We first briefly
describe this system and the way we adapted it by integrating our own set of corpus-based and lexical features.  We then
introduce the tagging models we have trained for 16 different languages using our adapted version of \melt. These models
are trained on the {\em Universal Dependencies} (v1.2) corpus set \citep{nivre15ud}, complemented by morphosyntactic
lexicons. We compare the accuracy of our models with the scores obtained by the CRF-based system \marmot
\citep{mueller13,muller15}, retrained on the same corpora and the same external morphosyntactic lexicons. We also compare
our results to those obtained by the best bidirectional LSTM models described by \citet{plank16}, which both make use
of Polyglot word vector representations published by \citet{alrfou13}. We will show that an optimised enrichment
of feature-based models with morphosyntactic lexicon results in significant accuracy gains. The macro-averaged accuracy
of our enriched \melt models is above that of enriched \marmot models and virtually identical to that of LSTMs enriched
with word vector representations. More precisely, per-language results indicate that lexicons provide more useful
information for languages with a high lexical variability (such as morphologically rich languages), whereas word vectors
are more informative for languages with a lower lexical variability (such as English).

\section{\protect\melt}
\label{sec:melt}

\melt \citep{denis12} is a tagging system based on maximum entropy Markov models (MEMM) \citep{ratnaparkhi96}, a
class of discriminative models that are suitable for sequence labelling \citep{ratnaparkhi96}.  The basic set of features
used by \melt is given in \citep{denis12}.  It is a superset of the feature sets used by \citet{ratnaparkhi96} and
\citet{toutanova00} and includes both {\em local standard features} (for example the current word itself and its
prefixes and suffixes of length 1 to 4) and {\em contextual standard features} (for example the tag just assigned to
the preceding word). In particular, with respect to Ratnaparkhi's feature set, \melt's basic feature set lifts the
restriction that local standard features used to analyse the internal composition of the current word should only apply
to rare words.

\begin{table*}[thp]
\centering\small
\begin{tabular}{rll}
\toprule
\multicolumn{3}{c}{\bf Local standard features}\\[1mm]
 & $\mathrm{wd}=w_i$ & $\wedge\ t_i=T$\\
$\forall k \in [1..4]$ & $\mathrm{pref}^k = w_i^1\ldots w_i^k$ & $\wedge\ t_i=T$\\
$\forall k \in [1..5]$ & $\mathrm{suff}^k = w_i^{n_i-k+1}\ldots w_i^{n_i}$ & $\wedge\ t_i=T$\\
 & $\mathrm{nb}=\mathrm{containsDigit}(w_i)$ & $\wedge\ t_i=T$\\
 & $\mathrm{hyph}=\mathrm{containsHyphen}(w_i)$ & $\wedge\ t_i=T$\\
 & $\mathrm{uc}=\mathrm{containsUppercase}(w_i)$ & $\wedge\ t_i=T$\\
 & $\mathrm{niuc}=(\mathrm{containsUppercase}(w_i) \wedge i>1)$ & $\wedge\ t_i=T$\\
 & $\mathrm{auc}=\mathrm{containsOnlyUppercase}(w_i)$ & $\wedge\ t_i=T$\\
\midrule
\multicolumn{3}{c}{\bf Contextual standard features}\\[1mm]
 & $\mathrm{wd}_{-2}=w_{i-2}$ & $\wedge\ t_i=T$\\
 & $\mathrm{wd}_{-1}=w_{i-1}$ & $\wedge\ t_i=T$\\
 & $\mathrm{wd}_{+1}=w_{i+1}$ & $\wedge\ t_i=T$\\
 & $\mathrm{wd}_{+2}=w_{i+2}$ & $\wedge\ t_i=T$\\
 & $\mathrm{swds}=w_{i-1}.w_{i+1}$ & $\wedge\ t_i=T$\\
$\forall k \in [1..3]$ & $\mathrm{pref}_{+1}^k = w_{i+1}^1\ldots w_i^k$ & $\wedge\ t_i=T$\\
$\forall k \in [1..3]$ & $\mathrm{suff}_{+1}^k = w_{i+1}^{n_i-k+1}\ldots w_i^{n_i}$ & $\wedge\ t_i=T$\\
 & $\mathrm{ptag}_{-2}=t_{i-2}$ & $\wedge\ t_i=T$\\
 & $\mathrm{ptag}_{-1}=t_{i-1}$ & $\wedge\ t_i=T$\\
 & $\mathrm{ptags}=t_{i-2}.t_{i-1}$ & $\wedge\ t_i=T$\\
\midrule
\multicolumn{3}{c}{\bf Local lexical features}\\[1mm]
$\mathrm{if\ lex}(w_i) = \{t\}$ & $\mathrm{lex_u}=t$ & $\wedge\ t_i=T$\\
$\mathrm{if}\ |\mathrm{lex}(w_i)| > 1, \forall t_j \in \mathrm{lex}(w_i)$ & $\mathrm{lex_{in}}=t_j$ & $\wedge\ t_i=T$\\
$\mathrm{if}\ |\mathrm{lex}(w_i)| > 1$ & $\mathrm{lex_{disj}}=\bigvee_{t_j\in\mathrm{lex}(w_i)} t_j$ & $\wedge\ t_i=T$\\
\midrule
\multicolumn{3}{c}{\bf Contextual lexical features}\\[1mm]
& $\mathrm{lex}_{+1}=\bigvee_{t_j\in\mathrm{lex}(w_{i+1})} t_j$ & $\wedge\ t_i=T$\\[2mm]
& $\mathrm{lex}_{+2}=\bigvee_{t_j\in\mathrm{lex}(w_{i+2})} t_j$ & $\wedge\ t_i=T$\\[2mm]
& $\mathrm{lex}_{+1.2}=\left(\bigvee_{t_j\in\mathrm{lex}(w_{i+1})} t_j\right).\left(\bigvee_{t_j\in\mathrm{lex}{w_{i+2}}} t_j\right)$ & $\wedge\ t_i=T$\\
\midrule
\multicolumn{3}{c}{\bf Contextual hybrid features}\\[1mm]
& $\mathrm{ptag}_{-1}.\mathrm{lex}_{+1}=t_{i-1}.(\bigvee_{t_j\in\mathrm{lex}(w_{i+1})} t_j)$ & $\wedge\ t_i=T$\\
\bottomrule
\end{tabular}
\caption{Feature set used by our \melt models. The current word is $w_i=w_i^1\ldots w_i^{n_i}$. Previously assigned tags
  for the two previous words are $t_{i-2}$ and $t_{i-1}$. The tag to be predicted for the current word is $t_i$,
  which can be assigned any tag $T$ in the tagset. The lex function applied to a word returns the set of all tags known to the
  lexicon for this word, or the singleton $\{\mathrm{\_unk\_\}}$ if the word is unknown to the lexicon. Boolean
  functions used by the local standard features have self-explanatory names.\label{tbl:features}}
\end{table*}

One of the advantages of feature-based models such as MEMMs and CRFs is that complementary information can be easily
added in the form of additional features. This was investigated for instance by \citet{kubler10}, whose
best-performing model for PoS tagging dialogues was obtained with a version of \melt extended with dialogue-specific
features. Yet the motivation of \melt's developers was first and foremost to investigate the best way to integrate
lexical information extracted from large-scale morphosyntactic lexical resources into their models, on top of the
training data \citep{denis12}. They showed that performances are better when this external lexical information is
integrated in the form of additional {\em lexical features} than when the external lexicon is used as constraints at
tagging time.\footnote{For instance by constraining the tagger in such a way that words known to the lexicon can only be
  associated with tags provided by the lexicon.} These lexical features can also be divided into {\em local lexical
  features} (for example the list of possible tags known to the external lexicon for the current word) and {\em
  contextual lexical features} (for example the list of possible tags known to the external lexicon for surrounding
words). In particular, lexical contextual features provide a means to model the right context of the current word, made
of words that have not yet been tagged by the system but for which the lexicon often provides a list of possible
tags. Moreover, tagging accuracy for out-of-vocabulary (OOV) words is improved, as a result of the fact that words
unknown to the training corpus might be known to the external lexicon.





\begin{table*}[tp]
\centering\small
\scalebox{0.92}{
\begin{tabular}{llrrl}
\toprule
{\bf Language} & {\bf Source Lexicon} & {\bf \#entries} & {\bf tagset size} & {\bf Reference} \\
\midrule
Bulgarian & Multext-EAST & 53056 & 12 & \citep{erjavec10}\\
Croatian & HML & 1360687 & 22 & \citep{oliver04}\\
Czech & Morfflex (parts) & 2094860 & 65 & \citep{hajic13}\\
Danish & STO & 566184 & 13 & \citep{braasch08}\\
English & EnLex & 472850 & 22 & \citep{sagot10lefff}\\
French & \lefff & 539278 & 25 & \citep{sagot10lefff}\\
German & DeLex & 465434 & 52 & \citep{sagot14delex}\\
Indonesian & Kateglo & 72217 & 118 & \url{https://github.com/ivanlanin/kateglo}\\
Italian & Morph-it! & 422756 & 31 & \citep{zanchetta05}\\
Norwegian & OrdBank & 679763 & 19 & \citep{hagen10}\\
Persian & PerLex & 511840 & 37 & \citep{sagot10perlex}\\
Polish & PolLex & 390370 & 28 & \citep{sagot07ltc}\\
Portuguese & LABEL-LEX$_\textrm{sw}$ & 971514 & 29 & \citep{ranchhod99}\\
Slovenian & SloLeks & 957525 & 25 & \citep{krek08}\\
Spanish & \leffe & 755858 & 34 & \citep{molinero09}\\
Swedish & Saldo & 747959 & 38 & \citep{borin08}\\\bottomrule
\end{tabular}
}
\caption{Information about the morphosyntactic lexicons used as external sources of lexical information in our \melt
  and \marmot models. The number of entries and tagset sizes refers to the morphosyntactic lexicons we extracted and
  used in our models, not to the original resources.\label{tbl:lexicons}}
\end{table*}

Despite a few experiments published with \melt on languages other than French \citep{denis12,leroux12,seddah13evalita},
the original feature set used by \melt (standard and lexical features) was designed and tested mostly on this language,
by building and evaluating tagging models on a variant of the French TreeBank. Since our goal was to carry out
experiments in a multilingual setting, we have decided to design our own set of features, using the standard \melt
features as a starting point. With respect to the original \melt feature set, we have added new ones, such as prefixes
and suffixes of the following word, as well as a hybrid contextual feature obtained by concatenating the tag predicted
for the preceding word and the tag(s) provided by the external lexicon for the following word.

In order to select the best performing feature set, we carried out a series of experiments using the multilingual
dataset provided during the SPMRL parsing shared task \citep{seddah13}. This included discarding useless or harmful
features and selecting the maximal length of the prefixes and suffixes to be used as features, both for the current word
and for the following word.\footnote{During these tuning experiments, we used development sets for comparing feature
  sets, without evaluating any of our models on test sets.}

We incorporated in \melt the best performing feature set, described in Table~\ref{tbl:features}. All models discussed in this
paper are based on this feature set.

\section{Datasets}
\label{sec:models}

\subsection{Corpora}

We carried out our experiments on the Universal Dependencies v1.2 treebanks \citep{nivre15ud}, hereafter UD1.2, from
which morphosyntactically annotated corpora can be trivially extracted. All UD1.2 corpora use a common tag set, the
17~{\em universal PoS tags},\footnote{\url{http://universaldependencies.org/u/pos/all.html}} which is an extension of
the tagset proposed by \citet{petrov12uts}.

As our goal is to study the impact of lexical information for PoS tagging, we have restricted our experiments to UD1.2
corpora that cover languages for which we have morphosyntactic lexicons at our disposal, and for which \citet{plank16}
provide results.\footnote{They discarded all corpora containing fewer than 60k tokens in the training set, maybe as a
  result of the sensitivy of LSMTs to training set size.} We considered UD1.2 corpora for the following 16 languages:
Bulgarian, Croatian, Czech, Danish, English, French, German, Indonesian, Italian, Norwegian, Persian, Polish,
Portuguese, Slovenian, Spanish and Swedish. Although this language list contains only one non-Indo-European
(Indonesian), four major Indo-European sub-families are represented (Germanic, Romance, Slavic, Indo-Iranian). Overall,
the 16 languages considered in our experiments are typologically, morphologically and syntactically fairly diverse.

\subsection{Lexicons}

We generate our external lexicons using the set of source lexicons listed in Table~\ref{tbl:lexicons}. Since external
lexical information is exploited via features, there is no need for the external lexicons and the annotated corpora to
use the same PoS inventory. Therefore, for each language, we simply extracted from the corresponding lexicon the PoS of
each word based on its morphological tags, by removing all information provided except for its coarsest-level
category.\footnote{However, for French, we used the morphosyntactic variant of the \lefff that is included in the \melt
  distribution, and which relies on a variant of the French TreeBank known as FTB-UC \citep{candito09iwpt}.} We also
added entries for punctuations when the source lexicons did not contain any.

We also performed experiments in which we retained the full original tags provided by the lexicons, with all
morphological features included. On average, results were slightly better than those presented in the paper, although
not statistically significantly. Moreover, the granularity of tag inventories in the lexicons is diverse, which makes it
difficult to draw general conclusions about results based on full tags. This is why we only report results based on
(coarse) PoS extracted from the original lexicons.

\section{Experiments and results}

\subsection{Baseline models}

In order to assess the respective contributions of external lexicons and word vector representations, we first compared
the results of the three above-mentioned systems when trained without such additional lexical
information. Table~\ref{tbl:results-baseline} provides the results of \melt and \marmot retrained on UD1.2 corpora,
together with the results publised on the same corpora by \citet{plank16}, using their best model not enhanced by
external word vector representations ---i.e.~the model they call $\vv{w}+\vv{c}$, which is a bidirectional LSTM that
combines both word and character embeddings.

\begin{table}[htp]
\centering\small
\scalebox{0.92}{
\begin{tabular}{llrrlrrrr}
\toprule
Model type& \multicolumn{1}{c}{\bf MEMM} & \multicolumn{1}{c}{\bf CRF} & \multicolumn{1}{c}{\bf bi-LSTM} \\
System& \multicolumn{1}{c}{\bf \melt} & \multicolumn{1}{c}{\bf \marmot} & \multicolumn{1}{c}{\bf Plank et al.} \\
\midrule
Bulgarian (bg)  &  97.75  &  97.64  &  {\bfseries 98.25}\\
Czech (cs)  &  98.01  &  {\bfseries 98.33}  &  97.93\\
Danish (da)  &  95.48  &  95.56  &  {\bfseries 95.94}\\
German (de)  &  92.74  &  92.85  &  {\bfseries 93.11}\\
English (en)  &  94.06  &  94.37  &  {\bfseries 94.61}\\
Spanish (es)  &  95.32  &  95.14  &  {\bfseries 95.34}\\
Persian (fa)  &  96.72  &  96.43  &  {\bfseries 96.89}\\
French (fr)  &  95.81  &  {\bfseries 96.13}  &  96.04\\
Croatian (hr)  &  95.08  &  95.15  &  {\bfseries 95.59}\\
Indonesian (id)  &  {\bfseries 93.74}  &  93.63  &  92.79\\
Italian (it)  &  97.44  &  {\bfseries 97.79}  &  97.64\\
Norwegian (no)  &  96.68  &  97.26  &  {\bfseries 97.77}\\
Polish (pl)  &  96.12  &  96.21  &  {\bfseries 96.62}\\
Portuguese (pt)  &  97.38  &  97.43  &  {\bfseries 97.48}\\
Slovene (sl)  &  96.05  &  96.23  &  {\bfseries 97.78}\\
Swedish (sv)  &  95.97  &  96.03  &  {\bfseries 96.30}\\
\midrule
Macro-avg. & 95.90 & 96.01 & {\bf 96.26}\\
\bottomrule
\end{tabular}
}
\caption{Overall accuracy (in \%) of baseline systems, i.e.~\melt and \marmot models trained without external lexicons, and
  Plank et al.'s (2016) $\vv{c}+\vv{w}$ models, which do not make use of Polyglot embeddings. Best scores are
  highlighted for each corpus. \label{tbl:results-baseline}}
\end{table}

These results show that Plank et al.'s (2016) bi-LSTM performs extremely well, surpassed by \marmot on only 3 out of 16
datasets (Czech, French and Italian), and by \melt only once (Indonesian).


\subsection{Models enriched with external lexical information}

Table~\ref{tbl:results-ud} provides the results of four systems enriched with lexical information. The feature-based
systems \melt and \marmot, respectively based on MEMMs and CRFs, are extended with the lexical information provided by
our morphosyntactic lexicons. This extension takes the form of additional features, as described in
Section~\ref{sec:melt} for \melt. The results reported by \citet{plank16} for their bidirectional LSTM when
initialised with Polyglot embeddings trained on full wikipedias are also included, together with their new system
FREQBIN, also initialised with Polyglot embeddings. FREQBIN trains bi-LSTMs to predict for each input word both a PoS
and a label that represents its log frequency in the training data. As they word it, ``the idea behind this model is to
make the representation predictive for frequency, which encourages the model not to share representations between common
and rare words, thus benefiting the handling of rare tokens.''

\begin{table*}[tp]
\centering\small
\scalebox{0.92}{
\begin{tabular}{lrrrrrrrr}
\toprule
Model type& \multicolumn{2}{c}{\bf MEMM+lexicon} & \multicolumn{2}{c}{\bf CRF+lexicon} & \multicolumn{2}{c}{\bf bi-LSTM+Polyglot} & \multicolumn{2}{c}{\bf FREQBIN+Polyglot} \\
System & \multicolumn{2}{c}{\bf \melt} & \multicolumn{2}{c}{\bf \marmot} & \multicolumn{4}{c}{\bf \citep{plank16}} \\
& overall & OOV & overall & OOV & overall & OOV & overall & OOV\\
\midrule
Bulgarian (bg)  &  98.15  &  93.95  &  98.05  &  93.06  &  {\bf 98.23}  &  87.40  &  97.97  &  {\bf 97.37}\\
Czech (cs)  &  {\bf 98.58}  &  94.83  &  98.48  &  93.68  &  98.02  &  89.02  &  97.89  &  {\bf 94.91}\\
Danish (da)  &  96.30  &  {\bf 92.32}  &  96.16  &  91.43  &  96.16  &  77.09  &  {\bf 96.35}  &  91.63\\
German (de)  &  93.43  &  88.08  &  93.10  &  87.21  &  {\bf 93.51}  &  81.95  &  93.38  &  {\bf 90.97}\\
English (en)  &  94.60  &  79.61  &  94.55  &  {\bf 79.99}  &  {\bf 95.17}  &  71.23  &  95.16  &  70.57\\
Spanish (es)  &  95.57  &  81.24  &  95.24  &  79.52  &  95.67  &  71.38  &  {\bf 95.74}  &  {\bf 98.22}\\
Persian (fa)  &  97.17  &  87.14  &  96.97  &  86.89  &  {\bf 97.60}  &  80.00  &  97.49  &  {\bf 96.54}\\
French (fr)  &  96.14  &  85.97  &  {\bf 96.34}  &  85.97  &  96.20  &  78.09  &  96.11  &  {\bf 92.13}\\
Croatian (hr)  &  96.70  &  93.01  &  96.19  &  91.23  &  96.27  &  84.62  &  {\bf 96.82}  &  {\bf 97.29}\\
Indonesian (id)  &  {\bf 93.83}  &  88.48  &  93.82  &  88.41  &  93.32  &  88.25  &  93.41  &  {\bf 94.70}\\
Italian (it)  &  97.82  &  91.98  &  {\bf 98.03}  &  91.82  &  97.90  &  83.59  &  97.95  &  {\bf 98.46}\\
Norwegian (no)  &  97.58  &  93.87  &  97.62  &  94.16  &  {\bf 98.06}  &  92.05  &  98.03  &  {\bf 97.78}\\
Polish (pl)  &  {\bf 97.77}  &  96.24  &  97.47  &  95.12  &  97.63  &  91.77  &  97.62  &  {\bf 99.35}\\
Portuguese (pt)  &  97.56  &  92.27  &  97.39  &  91.92  &  {\bf 97.94}  &  92.16  &  97.90  &  {\bf 96.87}\\
Slovene (sl)  &  {\bf 97.53}  &  96.50  &  97.23  &  94.89  &  96.97  &  80.48  &  96.84  &  {\bf 95.63}\\
Swedish (sv)  &  {\bf 96.90}  &  94.78  &  96.80  &  94.23  &  96.60  &  88.37  &  96.69  &  {\bf 96.02}\\
\midrule
 Macro-avg. {}  &  {\bf 96.60}  &  90.64  &  96.46  &  89.97  &  96.58  &  83.59  &  96.58  &  {\bf 94.28}\\
\bottomrule
\end{tabular}
}
\caption{Accuracy (in \%) of the feature-based systems \melt and \marmot as well as the two best LSTM-based systems by
  \citet{plank16} on UD1.2 datasets, which all use the 17 ``universal PoS tags''. \melt and \marmot models integrate the
  external lexicons listed in Table~\ref{tbl:lexicons}, whereas bidirectional LSTM-based systems rely on Polyglot word
  embeddings. Best scores overall and on OOV words are highlighted for each corpus. \label{tbl:results-ud}}
\end{table*}

\begin{figure}[htp]
\centering
\includegraphics[width=.5\columnwidth]{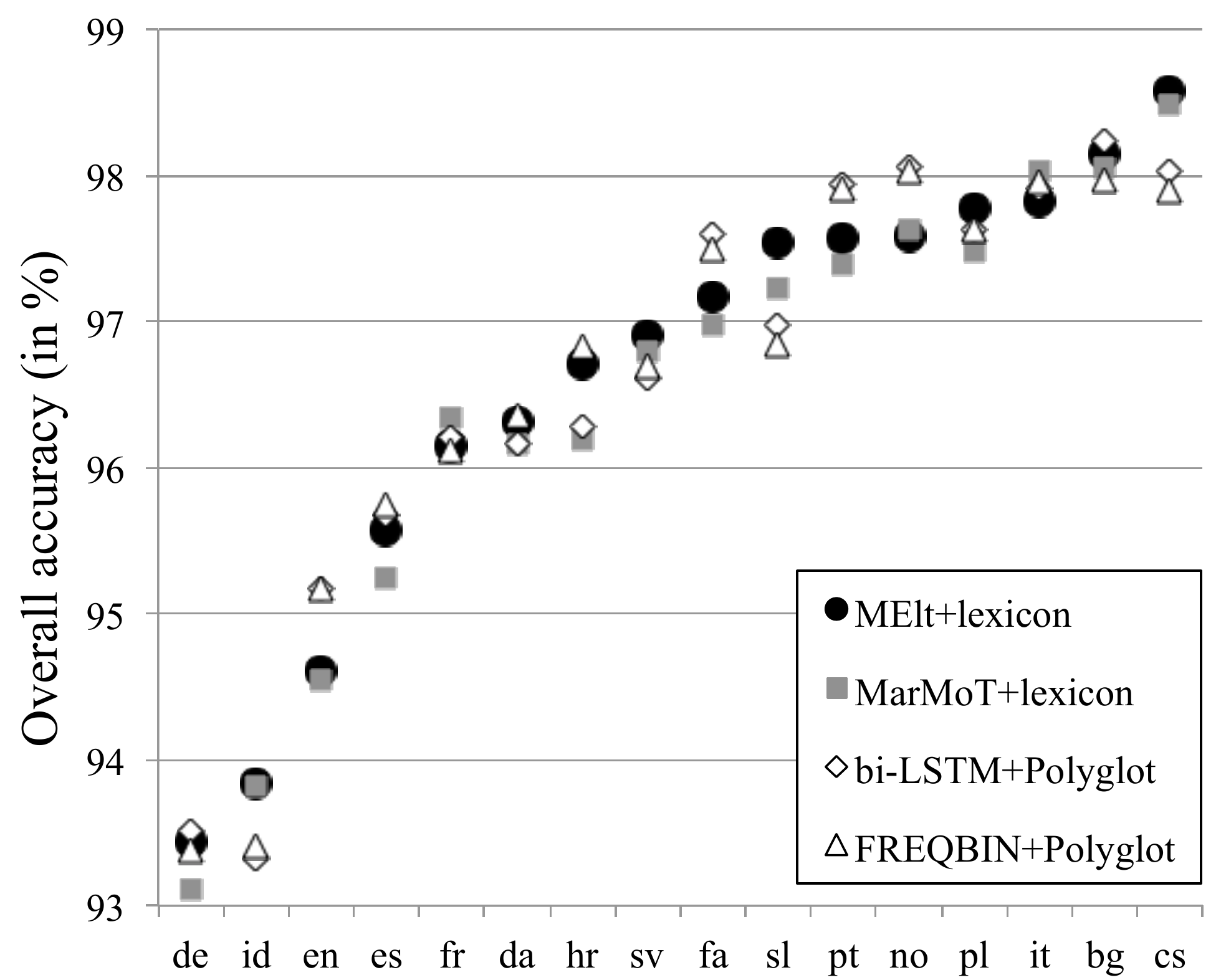}
\caption{Graphical visualisation of the overall tagging accuracies for all four types of enriched models. Detailed
  results are given in Table~\ref{tbl:results-ud}. Languages are sorted by increasing \melt's overall tagging
  scores.\label{fig:results}}
\end{figure}

\begin{figure}[htp]
\centering
\includegraphics[width=.5\columnwidth]{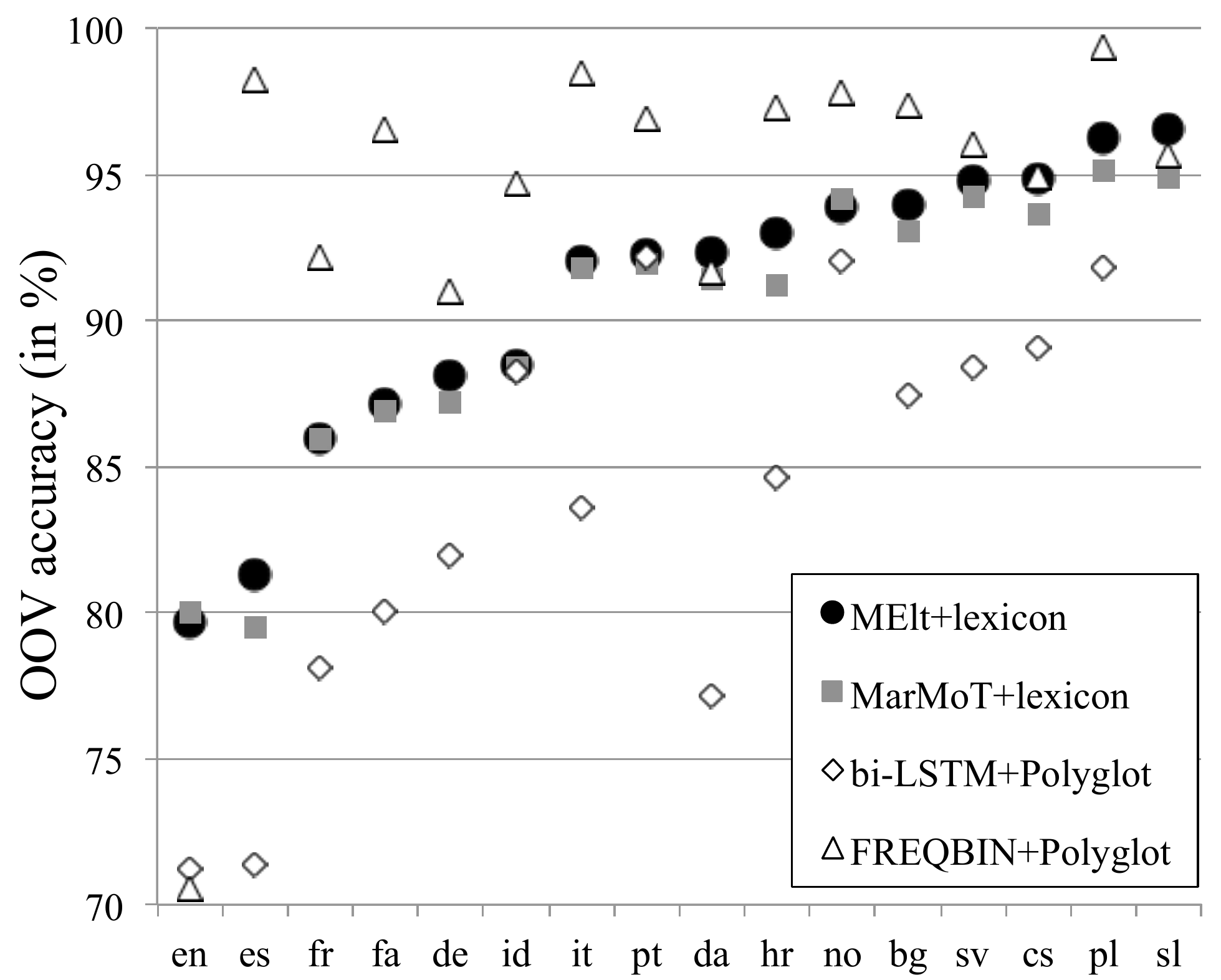}
\caption{Graphical visualisation of the OOV tagging accuracies for all types of models enriched with external lexicons.
  Detailed results are given in Table~\ref{tbl:results-ud}. Languages are sorted by increasing \melt's OOV tagging
  scores.\label{fig:resultsoov}}
\end{figure}

The results, which are also displayed in Figures~\ref{fig:results} and~\ref{fig:resultsoov}, show that all systems reach
very similar results on average, although discrepancies can be observed from one dataset to another, on which we shall
comment shortly. The best performing system in terms of macro-average is \melt (96.60\%). Both bi-LSTM systems reach the
same score (96.58\%), the difference with \melt's results being non significant, whereas \marmot is only 0.14\% behind
(96.46\%). Given the better baseline scores of the neural approaches, these results show that the benefit of using
external lexicons in the feature-based models \melt and \marmot are much higher than those using Polyglot word vector
representations as initialisations for bi-LSTMs.

Yet these very similar overall results reflect a different picture when focusing on OOV tagging accuracy. The best
models for OOV tagging accuracy are, by far, FREQBIN models, which are beaten by \marmot and by \melt only once each (on
English and Danish respectively). The comparison on OOV tagging between \melt and \marmot shows that \melt performs
better on average than \marmot, despite the fact that \marmot's baseline results were better than those reached by
\melt. This shows that the information provided by external morphosyntactic lexicons is better exploited by \melt's
lexical features than by those used by \marmot. On the other hand, the comparison of both bi-LSTM-based approaches
confirm that the FREQBIN models is better by over 10\% absolute on OOV tagging accuracy (94.28\% vs.~83.59\%), with 65\%
lower error rate.

\begin{figure}[htp]
\centering
\includegraphics[width=.5\columnwidth]{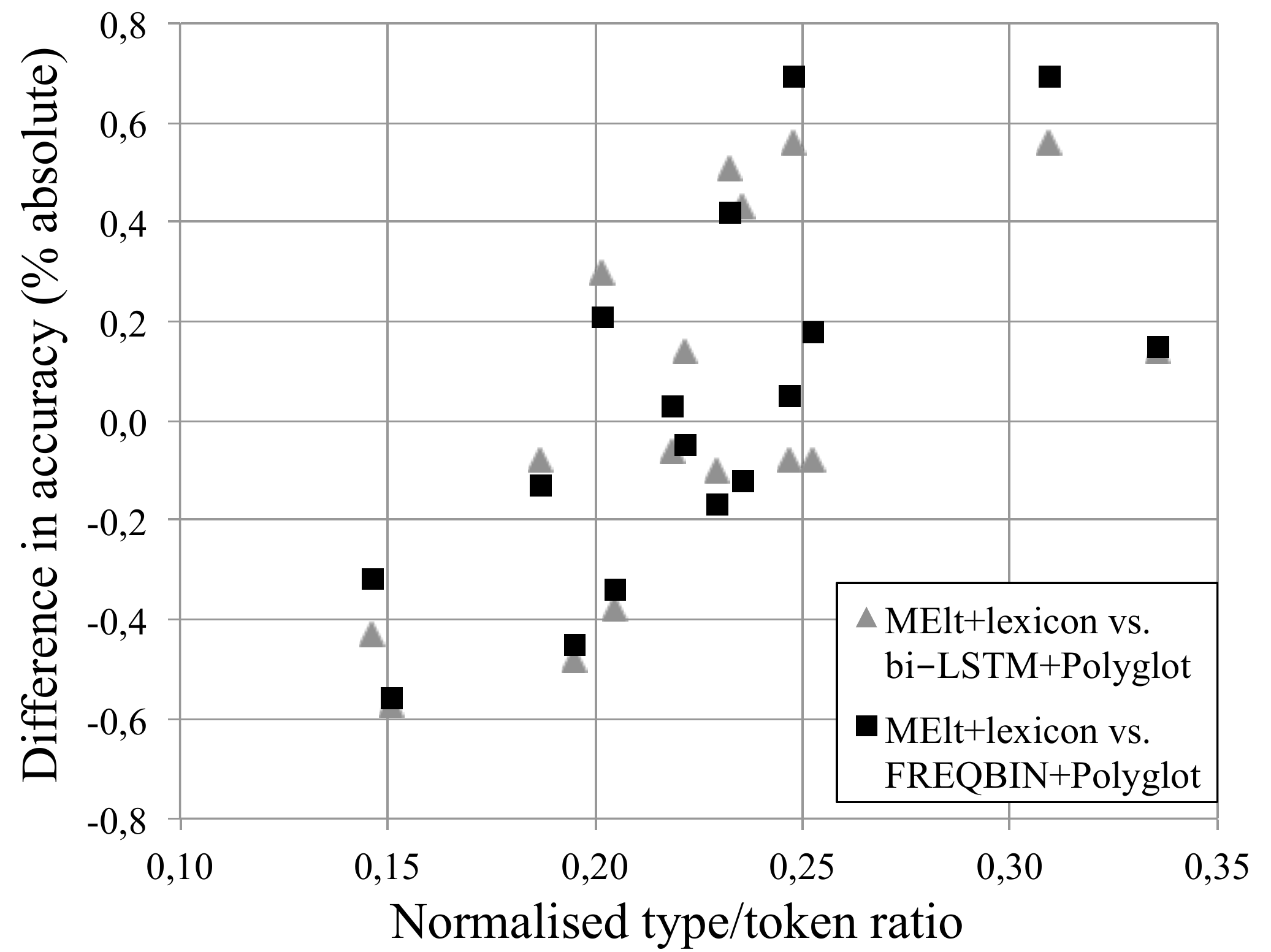}
\caption{Difference between the tagging accuracy of lexicon-enhanced \melt models and each of the two types of
  Polyglot-enhanced neural bi-LSTM models plotted against training sets' normalised type/token
  ratio.\label{fig:ttratio}}
\end{figure}

One of the important differences between the lexical information provided by an external lexicon and word vectors built
from raw corpora, apart from the very nature of the lexical information provided, is the coverage and accuracy of this
lexical information on rare words. All words in a morphosyntactic lexicon are associated with information of a same
granularity and quality, which is not the case with word representations such as provided by Polyglot. Models that take
advantage of external lexicons should therefore perform comparatively better on datasets containing a higher proportion
of rarer words, provided the lexicons' coverage is high. In order to confirm this intuition, we have used a lexical
richness metric based on the type/token ratio. Since this ratio is well-known for being sensitive to corpus length, we
normalised it by computing it over the 60,000 first tokens of each training set. When this normalised type/token ratio
is plotted against the difference between the results of \melt and both bi-LSTM-based models, the expected correlation
is clearly visible (see Figure~\ref{fig:ttratio}). This explains why \melt obtains better results on the morphologically
richer Slavic datasets (average normalised type/token ratio: 0.28, average accuracy difference: 0.32 compared to both
bi-LSTM+Polyglot and FREQBIN+Polyglot) and, at the other end of the spectrum, significantly worse results on the English
dataset (normalised type/token ratio: 0.15, average accuracy difference: -0.56 compared to bi-LSTM+Polyglot, -0.57
compared to FREQBIN+Polyglot).

\section{Conclusion}
\label{sec:conclusion}

Two main conclusions can be drawn from our comparative results. First, feature-based tagging models adequately enriched
with external morphosyntactic lexicons perform, on average, as well as bi-LSTMs enriched with word embeddings.
Per-language results show that the best accuracy levels are reached by feature-based models, and in particular by our
improved version of the MEMM-based system \melt, on datasets with high lexical variability (in short, for
morphologically rich languages), whereas neural-based results perform better on datatsets with lower lexical variability
(e.g.~for English).

We have only compared the contribution of morphosyntactic lexicons to feature-based models (MEMMs, CRFs) and that of
word vector representations to bi-LSTM-based models as reported by \citet{plank16}. As mentioned above, work on the
contribution of word vector representations to feature-based approaches has been carried out by
\citet{muller15}. However, the exploitation of existing morphosyntactic or morphological lexicons in neural models is
a less studied question.  Improvements over the state of the art might be achieved by integrating lexical information
both from an external lexicon and from word vector representations into tagging models.

In that regard, further work will be required to understand which class of models perform the best. An option would be
to integrate feature-based models such as a CRF with an LSTM-based layer, following recent proposals such as the one
proposed by \citet{lample16} for named entity recognition.
\tableofcontents

\bibliographystyle{apalike}
\bibliography{rr16melt}

\end{document}